\documentclass{article}
\usepackage{spconf,amsmath}

\usepackage[pdftex]{graphicx}
\usepackage[subrefformat=parens]{subcaption}
\usepackage{amsmath}
\usepackage{here}
\usepackage{cite}
\usepackage{comment}
\usepackage{multirow}

\usepackage{array}
\newcolumntype{L}[1]{>{\raggedright\let\newline\\\arraybackslash\hspace{0pt}}m{#1}}
\newcolumntype{C}[1]{>{\centering\let\newline\\\arraybackslash\hspace{0pt}}m{#1}}
\newcolumntype{R}[1]{>{\raggedleft\let\newline\\\arraybackslash\hspace{0pt}}m{#1}}

\title{SUPER-RESOLUTION USING CONVOLUTIONAL NEURAL NETWORKS WITHOUT ANY CHECKERBOARD ARTIFACTS}
\name{Yusuke Sugawara, Sayaka Shiota, and Hitoshi Kiya}
\address{Tokyo Metropolitan University, 6-6 Asahigaoka, Hino-shi, Tokyo, Japan}

\begin{document}

\makeatletter
\renewcommand\section{\@startsection{section}{1}{\z@}%
{0.5ex\@plus 0ex \@minus -.5ex}%
{0.5ex\@plus 0ex}%
{\normalfont\bfseries}
}
\renewcommand\subsection{\@startsection{subsection}{2}{\z@}%
{0.5ex\@plus 0ex \@minus -.5ex}%
{0.5ex\@plus 0ex}%
{\normalfont\bfseries}
}
\renewcommand\subsubsection{\@startsection{subsubsection}{3}{\z@}%
{0.5ex\@plus 0ex \@minus -.5ex}%
{0.5ex\@plus 0ex}%
{\normalfont\bfseries}
}
\makeatother

\renewcommand{\thesubsubsection}{\Alph{subsubsection}}

\setlength\floatsep{5pt}
\setlength\textfloatsep{5pt}
\setlength\intextsep{5pt}
\setlength\abovecaptionskip{5pt}
\setlength{\abovedisplayskip}{2pt} 
\setlength{\belowdisplayskip}{2pt} 

\ninept
\maketitle
\begin{abstract}
It is well-known that a number of excellent super-resolution (SR) methods using convolutional neural networks (CNNs) generate checkerboard artifacts.
A condition to avoid the checkerboard artifacts is proposed in this paper.
So far, checkerboard artifacts have been mainly studied for linear multirate systems,
but the condition to avoid checkerboard artifacts can not be applied to CNNs due to the non-linearity of CNNs.
We extend the avoiding condition for CNNs, and apply the proposed structure to some typical SR methods to confirm the effectiveness of the new scheme.
Experiment results demonstrate that the proposed structure can perfectly avoid to generate checkerboard artifacts under two loss conditions: mean square
error and perceptual loss, while keeping excellent properties that the SR methods have.
\end{abstract}
\section{Introduction}
\label{sec:intro}
\noindent This paper addresses the problem of checkerboard artifacts generated by some super-resolution (SR) methods using convolutional neural networks (CNNs).
SR methods using CNNs have been widely studying as one of single image SR techniques, and have superior performances \cite{SRCNN,SRCNN-Ex,VDSR,DRCN,DRRN}.
Moreover, in order to accelerate the processing speed, CNNs including upsampling layers such as deconvolution \cite{Deconv}
and sub-pixel convolution \cite{ESPCN} ones have been proposed \cite{ESPCN,FSRCNN,LapSRN,SRGAN,EnhanceNet,PSRnet}.
However, it is well-known that these SR methods generate periodic artifacts, referred to as checkerboard artifacts \cite{Perceptual_Loss}.
\par
In CNNs, it is well-known that checkerboard artifacts are generated by operations of deconvolution, sub-pixel convolution layers \cite{Checkerboard}.
To overcome these artifacts, smoothness constraint \cite{FlowNet}, post-processing \cite{Perceptual_Loss},
initialization scheme \cite{FreeSubPixel} and different upsampling layer designs \cite{Checkerboard,AdaptiveBilinear,PixelDeconv} have been proposed.
Most of them can not avoid checkerboard artifacts perfectly, although they reduce the artifacts.
Among them, Odena et al. \cite{Checkerboard} have demonstrated that
checkerboard artifacts can be perfectly avoided by using resize convolution layers instead of deconvolution ones.
However, the resize convolution layers can not be directly applied to upsampling layers such as deconvolution and sub-pixel convolution ones,
so this method needs not only large memory but also high computational costs.
\par
On the other hand, checkerboard artifacts have been studied to design linear multirate systems including filter banks and wavelets \cite{CB0,CB1,CB2,CB3}.
In addition, it is well-known that checkerboard artifacts are caused by the time-variant property of interpolators in multirate systems,
and the condition for avoiding these artifacts have been given \cite{CB0,CB1,CB2}.
However, the condition to avoid checkerboard artifacts for linear systems can not be applied to CNNs due to the non-linearity of CNNs.
\par
In this paper, we extend the avoiding condition for CNNs, and apply the proposed structure to SR methods using
deconvolution and sub-pixel convolution layers to confirm the effectiveness of the new scheme.
Experiment results demonstrate that the proposed structure can perfectly avoid to generate checkerboard artifacts under
two loss conditions: mean square error and perceptual loss, while keeping excellent properties that the SR methods have.
As a result, it is confirmed that the proposed structure allows us to offer efficient SR methods without any checkerboard artifacts.

\section{preparation}
\label{sec: preparation}
\noindent Conventional SR methods using CNNs and works related to checkerboard artifacts are reviewed, here.
\subsection{SR Methods using CNNs}
\label{subsec: SR methods using CNNs}
\begin{figure}[tb]
\centering
\centerline{\includegraphics[width=0.98\linewidth]{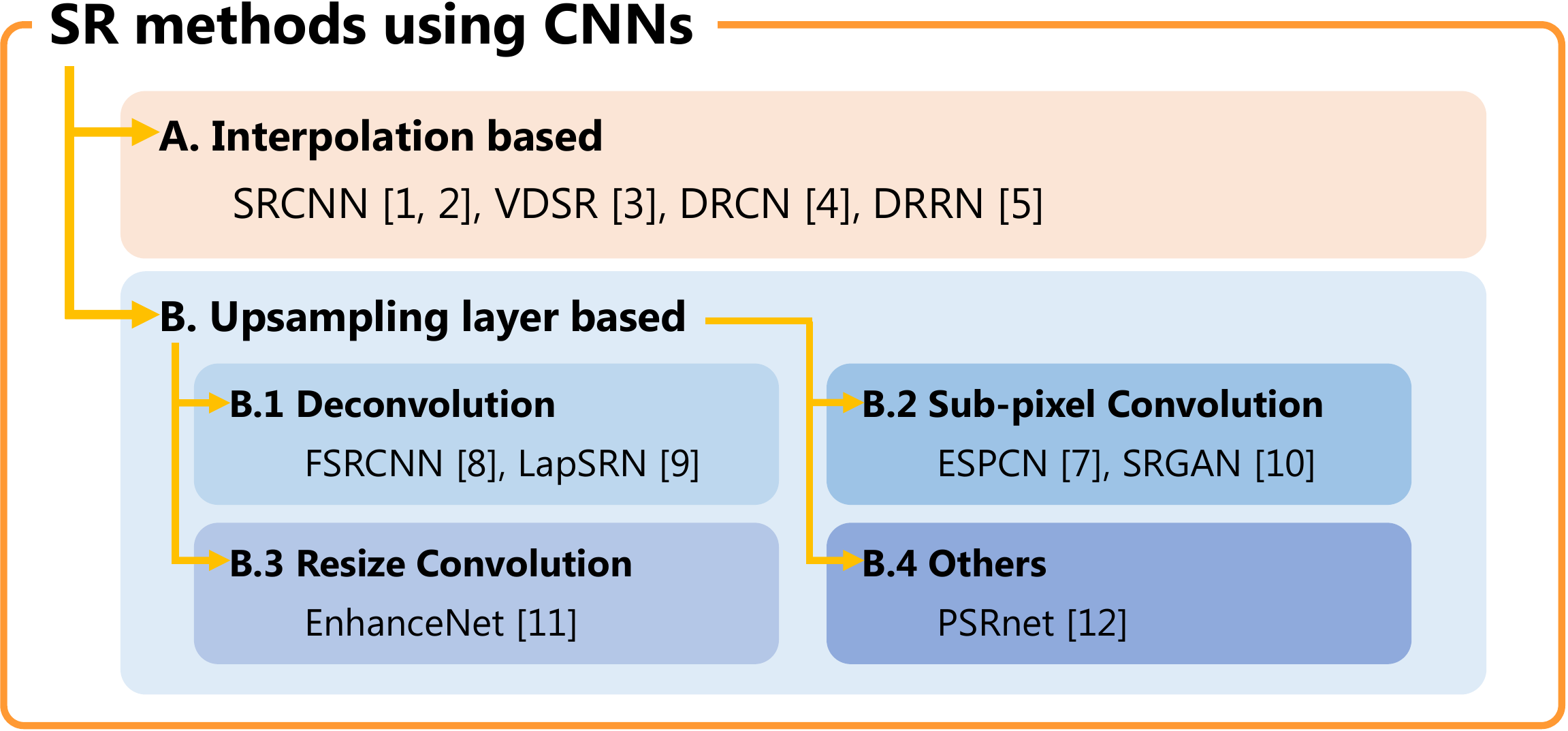}}
\caption{Classification of SR methods using CNNs}
\label{class}
\end{figure}

\noindent SR methods using CNNs are classified into two classes as shown in Fig. \ref{class}.
Interpolation based methods \cite{SRCNN,SRCNN-Ex,VDSR,DRCN,DRRN}, referred to as class A,
do not generate any checkerboard artifacts in CNNs, due to the use of an interpolated image as an input to a network.
In other words, CNNs in this class do not have any upsampling layers.
\par
On the other hand, when CNNs include upsampling layers, there is a possibility that the CNNs generate some checkerboard artifacts.
This class, called class B in this paper, have provided numerous excellent SR methods \cite{ESPCN,FSRCNN,LapSRN,SRGAN,EnhanceNet,PSRnet},
which can be executed faster than those in class A.
Class B is also classified into a number of sub-classes according to the type of upsampling layers.
This paper focuses on class B.
\par
\begin{figure}[tb]
\centering
\centerline{\includegraphics[width=0.8\linewidth]{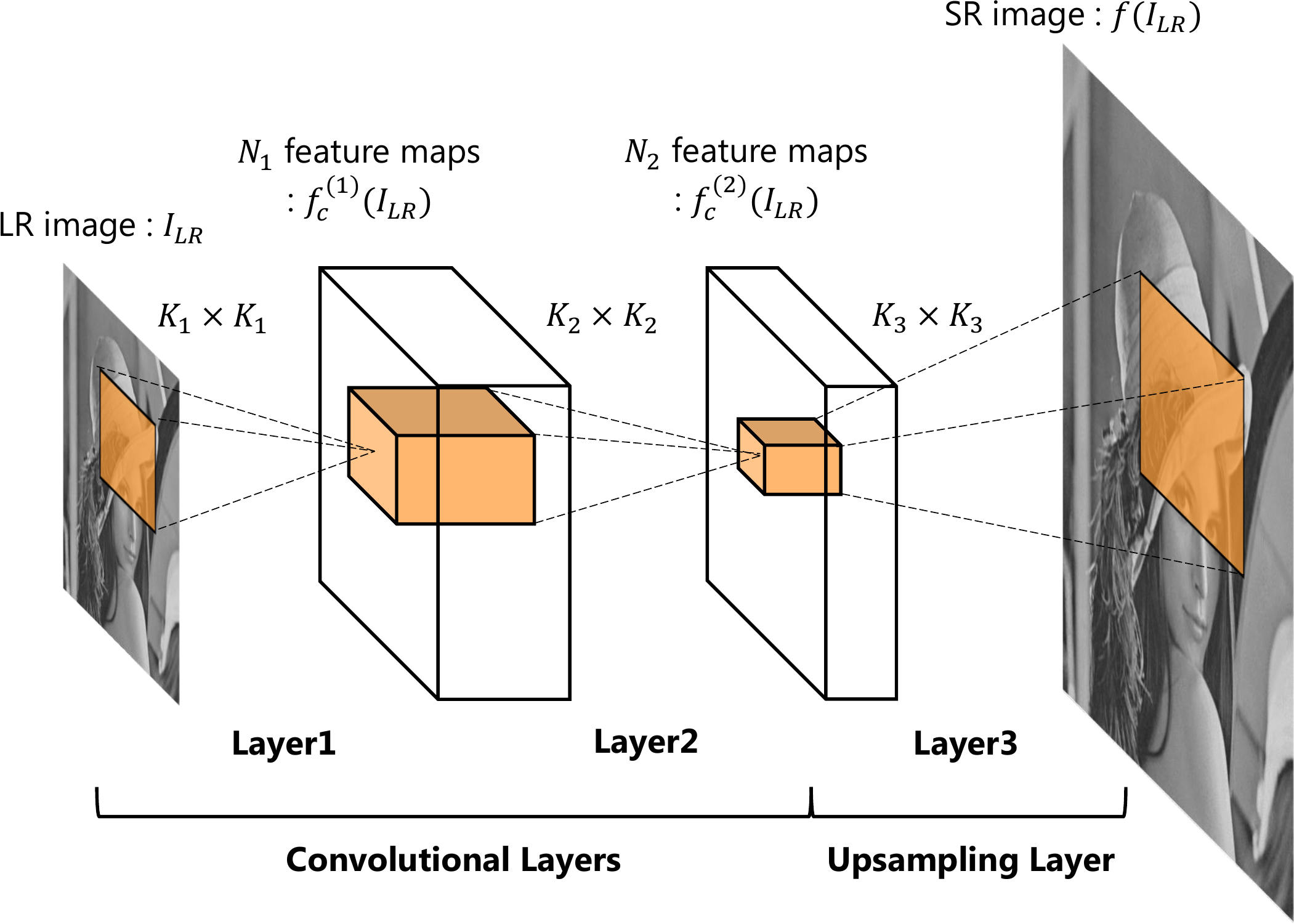}}
\caption{CNNs with an upsampling Layer}
\label{sr}
\end{figure}

CNNs are illustrated in Fig. \ref{sr} for an SR problem, as in \cite{ESPCN}, where the CNNs consist of two convolutional layers and one upsampling layer.
$I_{LR}$ and $f_c^{(l)}(I_{LR})$ are a low-resolution (LR) image and a $c$-th channel feature map at layer $l$, and $f(I_{LR})$ is an output of the network.
The two convolutional layers have learnable weights, biases, and ReLU \cite{ReLU} as an activation function, respectively,
where the weight at layer $l$ has $K_l \times K_l$ as a spatial size and $N_l$ as the number of feature maps.
\par
There are numerous algorithms for computing upsampling layers,
such as deconvolution, sub-pixel convolution and resize convolution ones, which are widely used as typical CNNs.
Besides, deconvolution \cite{Deconv}, sub-pixel convolution \cite{ESPCN} and resize convolution \cite{Checkerboard} layers
are well-known upsampling layers, respectively.

\subsection{Works Related to Checkerboard Artifacts}
\label{subsec: Works Related to Checkerboard Artifacts}
\noindent Checkerboard artifacts have been discussed to design multirate systems including filter banks and wavelets by researchers \cite{CB0,CB1,CB2,CB3}.
However, most of the works have been limited to in case of using linear systems, so they can not be directly applied to CNNs due to the non-linearity.
Some works related to checkerboard artifacts for linear systems are summarized, here.
\par
\begin{figure}[htb]
  \centering
  \begin{minipage}{\columnwidth}
    \centering
    \centerline{\includegraphics[width=0.8\linewidth]{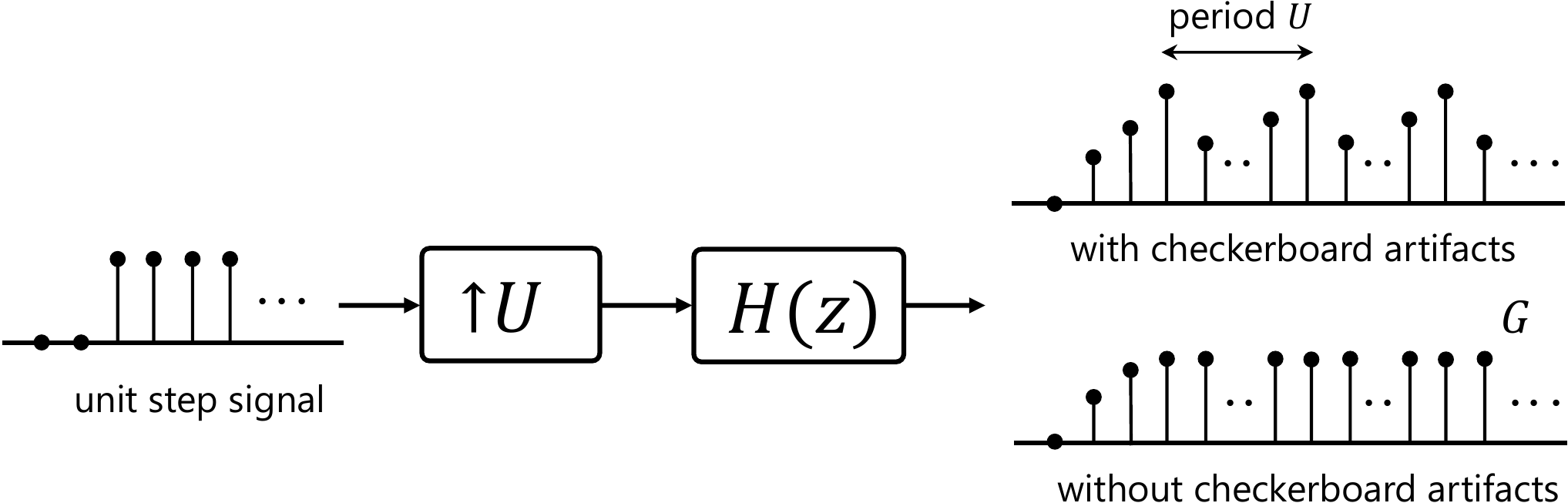}}
    \subcaption{General structure}
  \end{minipage}
  \vspace{-2mm}
  \begin{minipage}{\columnwidth}
    \centering
    \centerline{\includegraphics[width=0.9\linewidth]{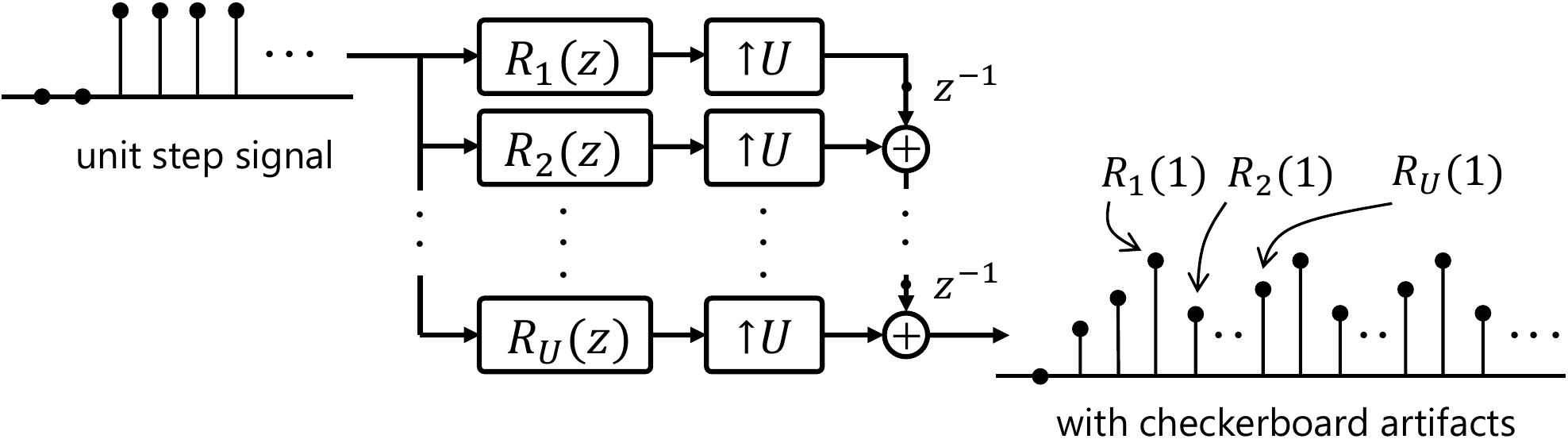}}
    \subcaption{Polyphase structure}
  \end{minipage}
  \caption{Linear interpolators with upscaling factor $U$}
\label{interpolator}
\end{figure}

It is known that linear interpolators which consist of up-samplers and linear time-invariant systems
cause checkerboard artifacts due to the periodic time-variant property \cite{CB0,CB1,CB2}.
Figure \ref{interpolator} illustrates a linear interpolator with an up-sampler $\uparrow U$ and a linear time-invariant system $H(z)$,
where positive integer $U$ is an upscaling factor and $H(z)$ is the $z$ transformation of an impulse response.
The interpolator in Fig. \ref{interpolator}(a) can be equivalently represented as a polyphase structure as shown in Fig. \ref{interpolator}(b).
The relationship between $H(z)$ and $R_i(z)$ is given by
\begin{equation}
\label{eq0}
H(z) = \sum_{i=1}^{U}R_{i}(z^{U})z^{-(U-i)},
\end{equation}
where $R_i(z)$ are often referred to as a polyphase filter of the filter $H(z)$.
\par
The necessary and sufficient condition for avoiding the checkerboard artifacts in the system is shown as
\begin{equation}
\label{eq1}
R_{1}(1) = R_{2}(1) = \cdots = R_{U}(1) = G.
\end{equation}
This condition means that all polyphase filters have the same DC value i.e. a constant $G$ \cite{CB0,CB1,CB2}.
Note that each DC value $R_i(1)$ corresponds to the steady-state value of the unit step response in each polyphase filter $R_i(z)$.
In addition, the condition eq.(\ref{eq1}) can be also expressed as
\begin{equation}
\label{eq2}
H(z) = P(z)H_0(z),
\end{equation}
where,
\begin{equation}
\label{eq3}
H_0(z) = \sum_{i=0}^{U-1} z^{-i},
\end{equation}
$H_0(z)$ and $P(z)$ are an interpolation kernel of the zero-order hold with factor $U$ and a time-invariant filter, respectively.
Therefore, the linear interpolator with factor $U$ does not generate any checkerboard artifacts, when $H(z)$ includes $H_0(z)$.
In the case without checkerboard artifacts, the step response of the linear system has a steady-state value $G$ as shown in Fig. \ref{interpolator}(a).
Meanwhile, the step response of the linear system has a periodic steady-state signal with the period of $U$,
such as $R_{1}(1)$, ..., $R_{U}(1)$, if eq.(\ref{eq2}) is not satisfied.

\section{proposed method}
\label{sec: proposed method}
\noindent CNNs are non-linear systems, so conventional works related to checkerboard artifacts can not be directly applied to CNNs.
A condition to avoid checkerboard artifacts in CNNs is proposed, here.
\subsection{CNNs with Upsampling Layers}
\label{subsec: Interpretation of upsampling layers using multirate systems}
\noindent We focus on upsampling layers in CNNs,
for which there are numerous algorithms such as deconvolution \cite{Deconv}, sub-pixel convolution \cite{ESPCN} and resize convolution \cite{Checkerboard}.
For simplicity, one-dimensional CNNs will be considered in the following discussion.
\par
\begin{figure}[tb]
  \centering
  \begin{minipage}{\columnwidth}
    \centering
    \centerline{\includegraphics[width=0.8\linewidth]{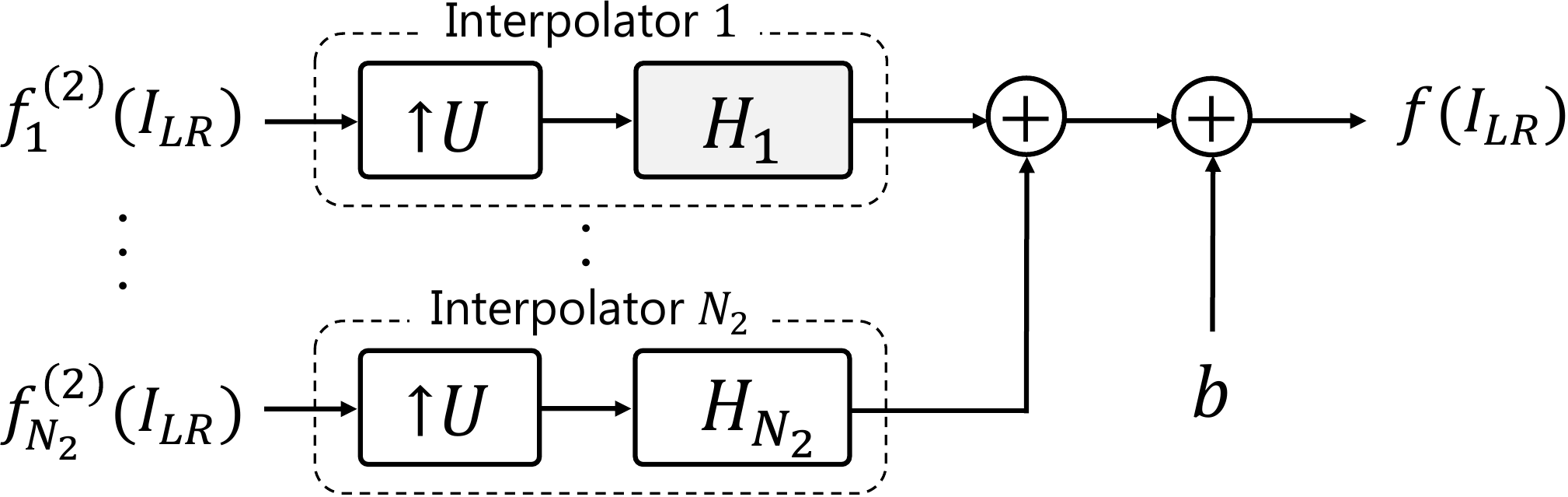}}
    \subcaption{General structure}
  \end{minipage}
  \vspace{-2mm}
  \begin{minipage}{\columnwidth}
    \centering
    \centerline{\includegraphics[width=0.8\linewidth]{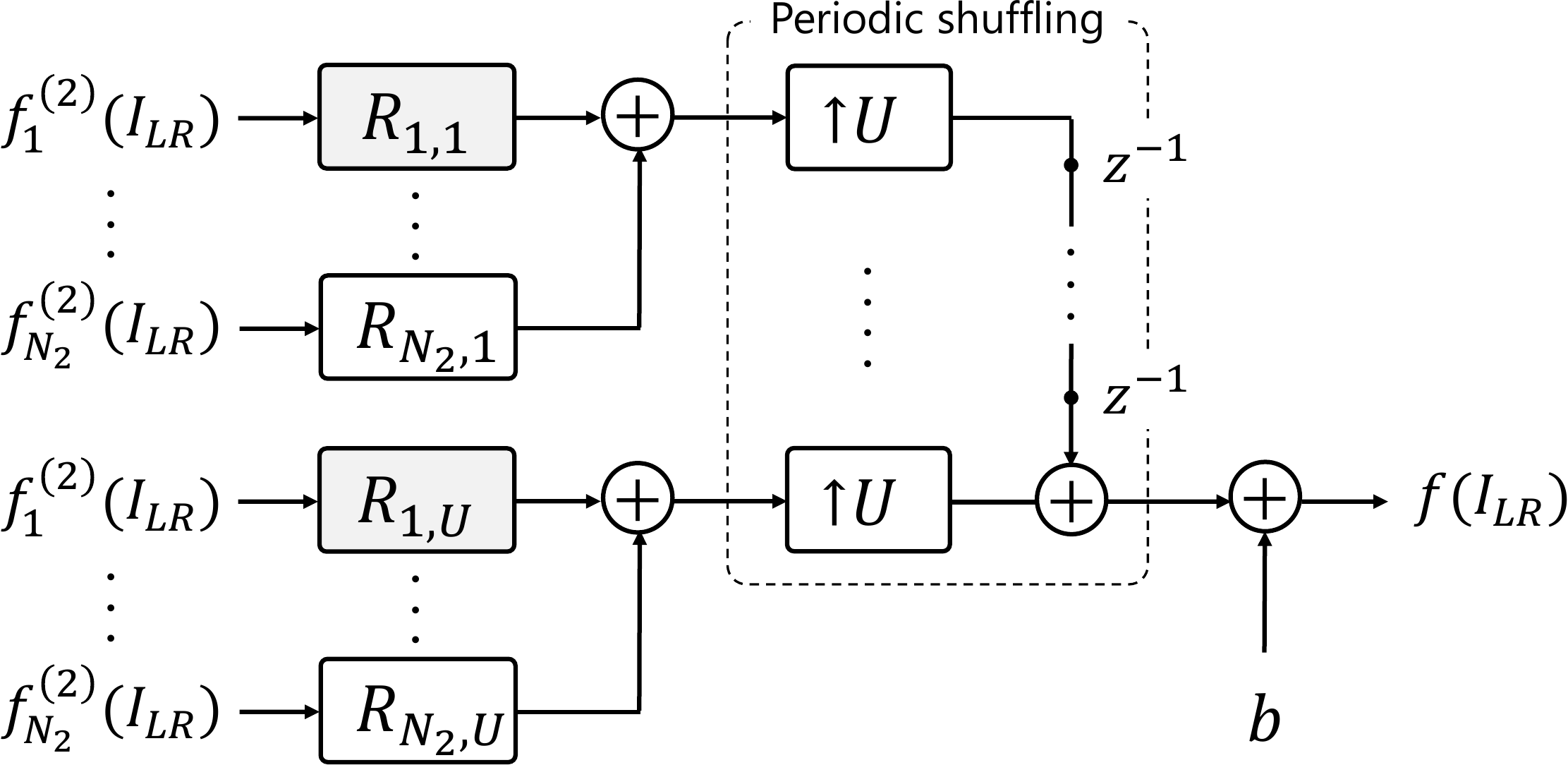}}
    \subcaption{Polyphase structure}
  \end{minipage}
  \caption{Deconvolution layer \cite{Deconv}}
  \label{deconv}
\end{figure}

It is well-known that deconvolution layers with non-unit strides cause checkerboard artifacts \cite{Checkerboard}.
Figure \ref{deconv} illustrates a system representation of deconvolution layers \cite{Deconv} which consist of some interpolators,
where $H_{c}$ and $b$ are a weight and a bias in which $c$ is a channel index, respectively.
The deconvolution layer in Fig. \ref{deconv}(a) can be equivalently represented as a polyphase structure in Fig. \ref{deconv}(b),
where $R_{c,n}$ is a polyphase filter of the filter $H_{c}$ in which $n$ is a filter index.
This is a non-linear system due to the bias $b$.
\par
Figure \ref{sub-pixel} illustrates a representation of sub-pixel convolution layers \cite{ESPCN},
where $R_{c,n}$ and $b_{n}$ are a weight and a bias, and $f_n^{\prime}(I_{LR})$ is an intermediate feature map in channel $n$.
Compared Fig.\ref{deconv}(b) with Fig.\ref{sub-pixel}, we can see that the polyphase structure in Fig. \ref{deconv}(b) is
a special case of sub-pixel convolution layers in Fig. \ref{sub-pixel}.
In other words, Fig. \ref{sub-pixel} is reduced to Fig. \ref{deconv}(b), when satisfying $b_{1}=b_{2}=...=b_{U}$.
Therefore, we will focus on sub-pixel convolution layers as the general case of upsampling layers to discuss checkerboard artifacts in CNNs.

\subsection{Checkerboard Artifacts in CNNs}
\label{subsec: Definition of Checkerboard Artifacts in CNNs}
\begin{figure}[tb]
\centering
\centerline{\includegraphics[width=0.8\linewidth]{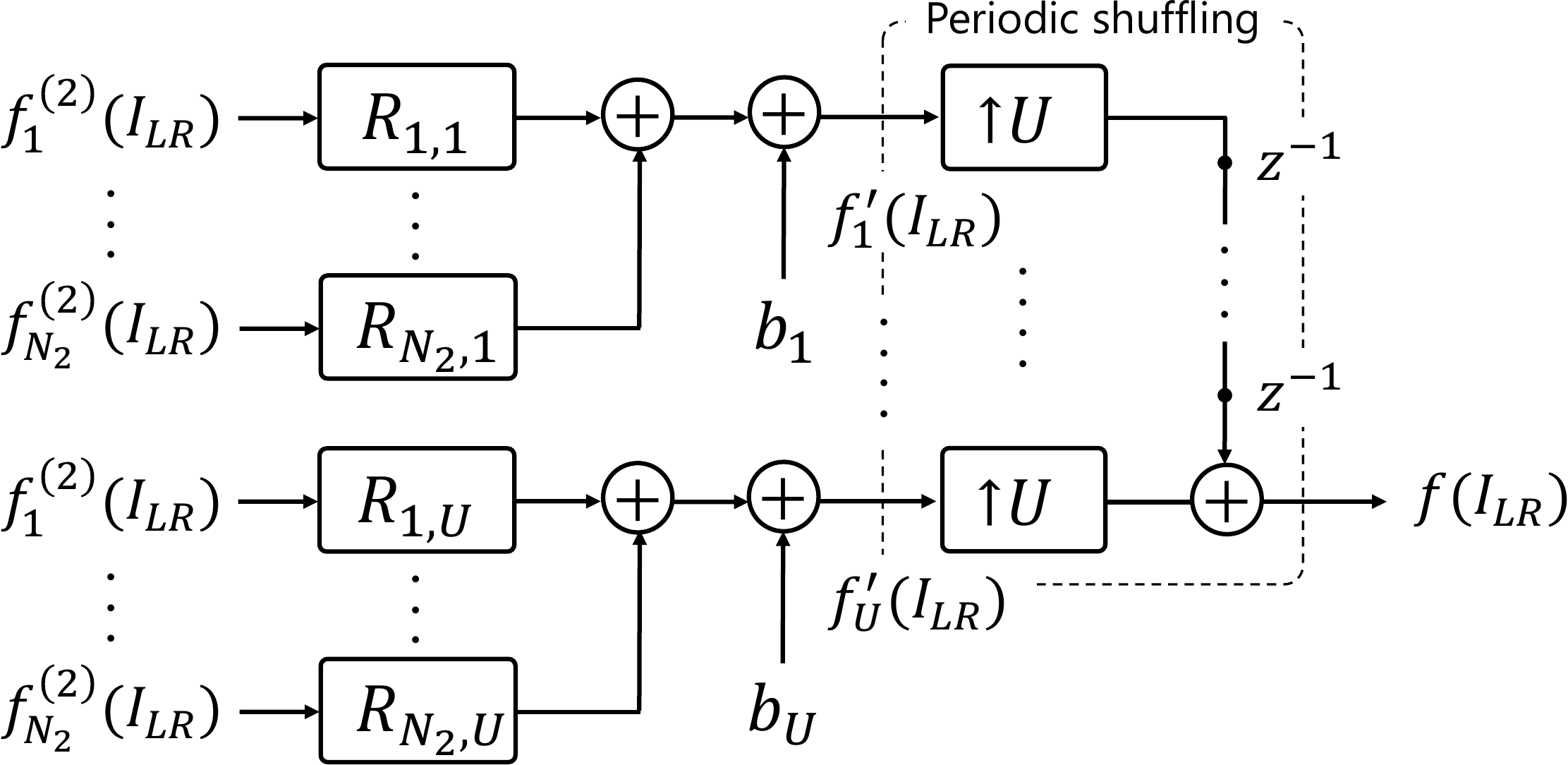}}
\caption{Sub-pixel convolution layer \cite{ESPCN}}
\label{sub-pixel}
\end{figure}

\noindent Let us consider the unit step response in CNNs.
In Fig. \ref{sr}, when the input $I_{LR}$ is the unit step signal $I_{step}$,
the steady-state value of the $c$-th channel feature map in layer $2$ is given as
\begin{equation}
\label{eq4}
\hat{f}_c^{(2)}(I_{step}) = A_c,
\end{equation}
where $A_c$ is a positive constant value, which is decided by filters, biases and ReLU.
Therefore, from Fig. \ref{sub-pixel}, the steady-state value of the $n$-th channel intermediate feature map is given by, for sub-pixel convolution layers,
\begin{equation}
\label{eq5}
\hat{f}_n^{\prime}(I_{step}) = \,\sum_{c=1}^{N_2} \, A_c \overline{R}_{c,n}\, + b_n,
\end{equation}
where $\overline{R}_{c,n}$ is the DC value of the filter $R_{c,n}$.
\par
Generally, the condition,
\begin{equation}
\label{eq5-}
\hat{f}_1^{\prime}(I_{step}) = \hat{f}_2^{\prime}(I_{step}) = ... = \hat{f}_U^{\prime}(I_{step}),
\end{equation}
is not satisfied, so the unit step response $f(I_{step})$ has a periodic steady-state signal with the period of $U$.
To avoid checkerboard artifacts, eq.(\ref{eq5-}) has to be satisfied, as well as for linear multirate systems.

\subsection{Upsampling Layers without Checkerboard Artifacts}
\label{subsec: Upsampling Layers without Checkerboard Artifacts}
\noindent To avoid checkerboard artifacts, CNNs must have the non-periodic steady-state value of the unit step response.
From eq.(\ref{eq5}), eq.(\ref{eq5-}) is satisfied, if
\begin{equation}
\label{eq6}
\overline{R}_{c,1} = \overline{R}_{c,2} = \cdots = \overline{R}_{c,U}, \, c = 1, 2, ..., N_{2}
\end{equation}
\begin{equation}
\label{eq7}
b_{1} = b_{2} = \cdots = b_{U},
\end{equation}
Note that, in this case,
\begin{equation}
\label{eq8}
\hat{f}_1^{\prime}(K \cdot I_{step}) = \hat{f}_2^{\prime}(K \cdot I_{step}) = ... = \hat{f}_U^{\prime}(K \cdot I_{step}),
\end{equation}
is also satisfied as for linear systems, where $K$ is an arbitrary constant value.
However, even when each filter $H_c$ in Fig.\ref{sub-pixel} satisfies eq.(\ref{eq2}),
eq.(\ref{eq7}) is not met, but eq.(\ref{eq6}) is met.
Therefore, we have to seek for a new insight to avoid checkerboard artifacts in CNNs.
\par
\begin{figure}[tb]
\centering
\centerline{\includegraphics[width=0.9\linewidth]{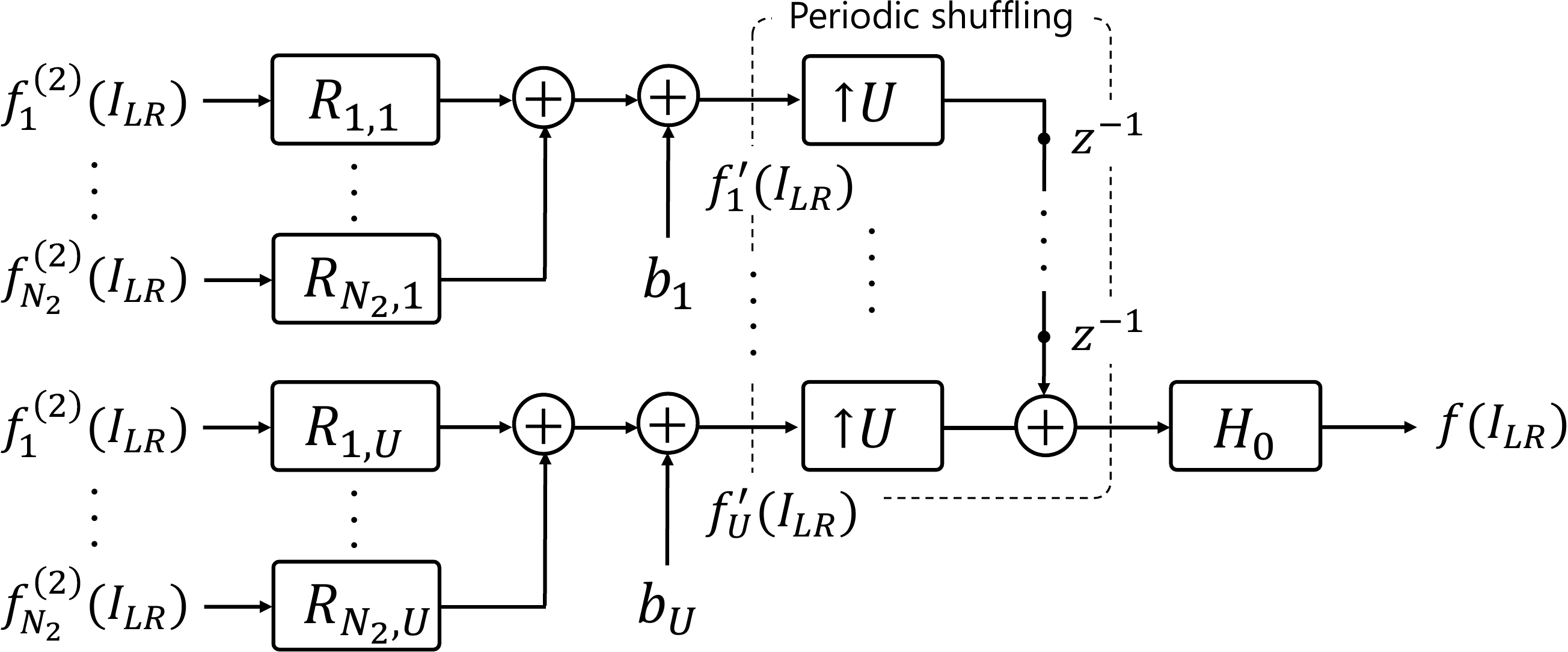}}
\caption{Proposed upsampling layer structure without checkerboard artifacts}
\label{Pro}
\end{figure}

In this paper, we propose to add the kernel of the zero-order hold with factor $U$, i.e. $H_0$ in eq.(\ref{eq3}), after upsampling layers as shown in Fig. \ref{Pro}.
In this structure, the output signal from $H_0$ can be a constant value, even when an arbitrary periodic signal is inputted to $H_0$.
As a result, Fig. \ref{Pro} can satisfy eq.(\ref{eq5-}).
\par
There are three approaches to use $H_0$ in CNNs by the difference in training CNNs as follows.
\subsubsection{Training CNNs without $H_0$}
\noindent The simplest approach for avoiding checkerboard artifacts is to add $H_0$ to CNNs after training the CNNs.
This approach allows us to perfectly avoid checkerboard artifacts generated by a pre-trained model.
\subsubsection{Training CNNs with $H_0$}
\noindent In approach B, $H_0$ is added to CNNs before training the CNNs, and then the CNNs with $H_0$ are trained.
This approach also allows us to perfectly avoid checkerboard artifacts as well as for approach A.
Moreover, this approach provides higher quality images than those of approach A.
\subsubsection{Training CNNs with $H_0$ inside upsampling layers}
\noindent Approach C is applicable to only deconvolution layers,
although approaches A and B are available for both of deconvolution layers and sub-pixel convolution ones.
Deconvolution layers always satisfy eq.(\ref{eq7}), so eq.(\ref{eq6}) only has to be considered.
Therefore, CNNs do not generate any checkerboard artifacts when each filter $H_c$ in Fig.5 satisfies eq.(\ref{eq2}).
In approach C, checkerboard artifacts are avoided by convolving each filter $H_c$ with the kernel $H_0$ inside upsampling layers.

\begin{figure*}[tb]
\centering
\centerline{\includegraphics[width=0.9\linewidth]{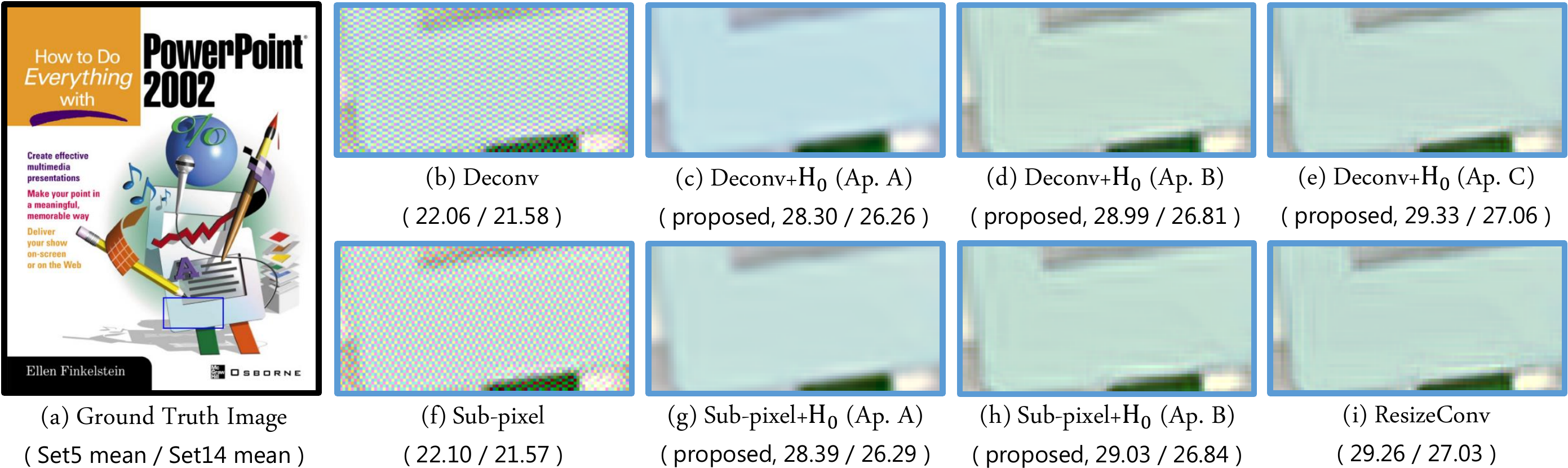}}
\caption{Experimental results of super-resolution under perceptual loss (PSNR(dB))}
\label{VGG}
\end{figure*}
\begin{figure*}[tb]
\centering
\centerline{\includegraphics[width=0.9\linewidth]{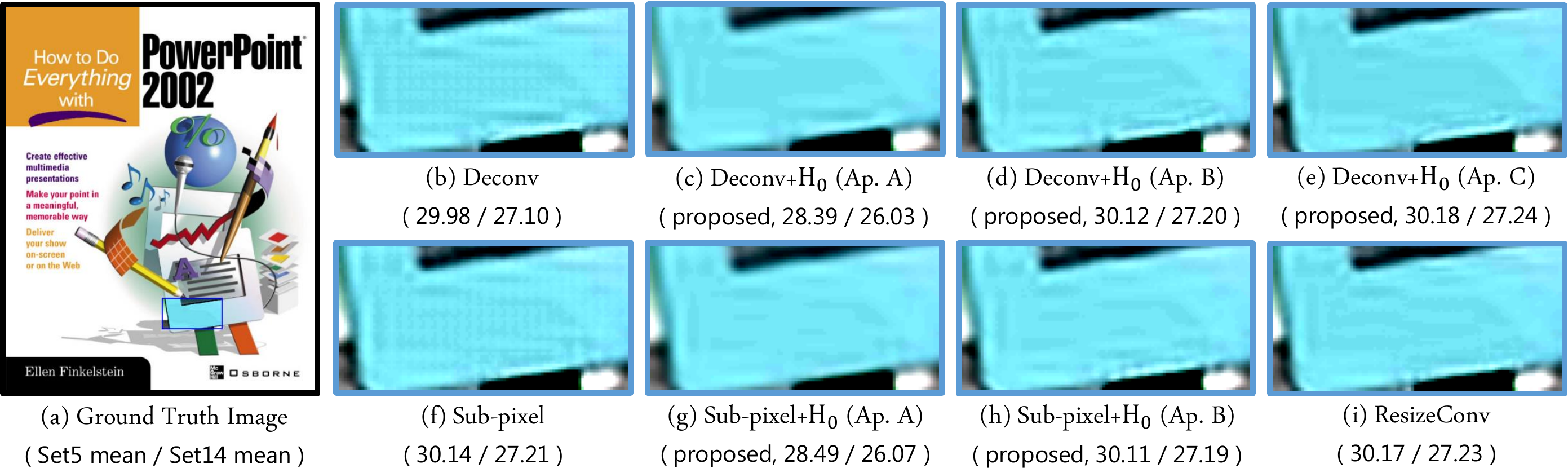}}
\caption{Experimental results of super-resolution under MSE loss (PSNR(dB))}
\label{MSE}
\end{figure*}

\section{experiments and results}
\label{sec: experiments and results}
\noindent The proposed structure without checkerboard artifacts was applied to
the SR methods using deconvolution and sub-pixel convolution layers to demonstrate the effectiveness.
CNNs in the experiments were carried out under two loss functions: mean squared error (MSE) and perceptual loss.
\subsection{Datasets for Training and Testing}
\label{subsec: Datasets for Training and Testing}
\noindent We employed 91-image set from Yang et al. \cite{91image} as our training dataset.
In addition, the same data augmentation (rotation and downscaling) as in \cite{FSRCNN} was used.
As a result, the training dataset consisting of 1820 images was created for our experiments.
Besides, we used two datasets, Set5 \cite{Set5} and Set14 \cite{Set14}, which are often used for benchmark, as test datasets.
\par
To prepare a training set, we first downscaled the ground truth images $I_{HR}$
with a bicubic kernel to create the LR images $I_{LR}$, where the factor $U=4$ was used.
The ground truth images $I_{HR}$ were cropped into $72 \times 72$ pixel patches
and the LR images were also cropped $18 \times 18$ pixel ones, where the total number of extracted patches was $8,000$.
In the experiments, the luminance channel (Y) of images was used for the MSE loss,
although the three channels (RGB) of images were used for the perceptual loss.

\subsection{Training Details}
\label{subsec: Training Details}
\noindent Table \ref{CNNs} illustrates CNNs used in the experiments, which were carried out based on CNNs in Fig. \ref{sr}.
For other two layers in Fig. \ref{sr}, we set $(K_1,N_1)=(5,64)$, $(K_2,N_2)=(3,32)$ as in \cite{ESPCN}.
In addition, the training of all networks was carried out to minimize the mean squared error $\frac{1}{2}\|I_{HR}-f(I_{LR})\|^2$
and the perceptual loss $\frac{1}{2}\|\phi(I_{HR})-\phi(f(I_{LR}))\|^2$ averaged over the training set, respectively,
where $\phi$ calculates feature maps at the fourth layer of the pre-trained VGG-16 model as in \cite{Perceptual_Loss}.
It is well-known that the perceptual loss results in sharper SR images despite lower PSNR values,
and generates checkerboard artifacts more frequently than under the MSE loss.
\begin{table}[tb]
\caption{CNNs used in the experiments}
\scalebox{0.73}{
  \begin{tabular}{l|l|c}
    Network Name & Upsampling Layer & $K_3 \times K_3$ \\ \hline
    \textbf{Deconv} & Deconvolution \cite{Deconv} & $9 \times 9$ \\
    \textbf{Sub-pixel} & Sub-pixel Convolution \cite{ESPCN} & $3 \times 3$ \\
    \textbf{ResizeConv} & Resize Convolution \cite{Checkerboard} & $9 \times 9$ \\
    \textbf{Deconv+$\rm H_0$} & Deconvolution with $H_0$ ( Approach A or B ) & $9 \times 9$ \\
    \textbf{Deconv+$\rm H_0$ (Ap. C)} & Deconvolution with $H_0$ ( Approach C ) & $9 \times 9$ \\
    \textbf{Sub-pixel+$\rm H_0$} & Sub-pixel Convolution with $H_0$ ( Approach A or B ) & $3 \times 3$ \\
  \end{tabular}}
  \label{CNNs}
\end{table}
\par
For training, Adam \cite{Adam} with $\beta_1=0.9, \beta_2=0.999$ was employed as an optimizer.
Besides, we set the batch size to $4$ and the learning rate to $0.0001$.
The weights were initialized with the method described in He et al. \cite{He}.
We trained all models for $200$K iterations.
All models were implemented by using the tensorflow framework \cite{Tensorflow}.

\subsection{Experimental Results}
\noindent Figure \ref{VGG} shows examples of SR images generated under the perceptual loss, where mean PSNR values for each dataset are also illustrated.
In this figure, (b) and (f) include checkerboard artifacts, although (c), (d), (e), (g), (h) and (i) do not include any ones.
Moreover, it is shown that the quality of SR images was significantly improved by avoiding checkerboard artifacts.
Approach B and C also provided better quality images than approach A.
In Fig. \ref{MSE}, (b) and (f) also include checkerboard artifacts as well as in Fig. \ref{VGG}, although the distortion is not so large,
compared to under the perceptual loss.
Note that ResizeConv does not generate any checkerboard artifacts, because it uses a pre-defined interpolation like in \cite{SRCNN}.
\par
Table \ref{Time} illustrates the average executing time when each CNNs were carried out 10 times for some images in Set14.
ResizeConv needs the highest computational cost in this table, although it does not generate any checkerboard artifacts.
From this table, the proposed structures have much lower computational costs than with resize convolution layers.
Note that the result was tested on PC with a 3.30 GHz CPU and the main memory of 16GB.
\begin{table}[tb]
  \begin{center}
    \caption{Execution time of super-resolution (sec)}
    \label{Time}
    \scalebox{0.75}{
    \begin{tabular}{ C{2cm} || C{2.4cm} | C{2.4cm} | C{2.4cm} } \hline
     Resolution & \multirow{2}{*}{Deconv} & Deconv+$\rm H_0$   & Deconv+$\rm H_0$ \\
    of Input Image &        & ( Ap. A or B ) & ( Ap. C ) \\ \hline\hline
     $69\times69$      & 0.00871 & 0.0115 & 0.0100 \\
     $125\times90\;\,$ & 0.0185  & 0.0270 & 0.0227 \\
     $128\times128$    & 0.0244  & 0.0348 & 0.0295 \\
     $132\times164$    & 0.0291  & 0.0393 & 0.0377 \\
     $180\times144$    & 0.0343  & 0.0476 & 0.0421 \\ \hline \noalign{\vskip2.5mm} \hline
     Resolution & \multirow{2}{*}{Sub-pixel} & Sub-pixel+$\rm H_0$ & \multirow{2}{*}{ResizeConv} \\
    of Input Image &           & ( Ap. A or B )  &  \\ \hline\hline
     $69\times69$      & 0.0159  & 0.0242 & 0.107 \\
     $125\times90\;\,$ & 0.0398  & 0.0558 & 0.224 \\
     $128\times128$    & 0.0437  & 0.0619 & 0.299 \\
     $132\times164$    & 0.0696  & 0.0806 & 0.383 \\
     $180\times144$    & 0.0647  & 0.102  & 0.450 \\ \hline
    \end{tabular}
    }
  \end{center}
\end{table}

\section{conclusion}
\label{conclusion}
\noindent This paper addressed a condition to avoid checkerboard artifacts in CNNs including upsampling layers.
The proposed structure can be applied to both of deconvolution layers and sub-pixel convolution ones.
The experimental results demonstrated that the proposed structure can perfectly avoid to generate checkerboard artifacts under two loss functions:
mean squared error and perceptual loss, while keeping excellent properties that the SR methods have.
As a result, the proposed structure allows us to offer efficient SR methods without any checkerboard artifacts.
The proposed structure will be also useful for various computer vision tasks
such as semantic segmentation, image synthesis and image generation.

\newpage
\bibliographystyle{IEEEbib}
\bibliography{refs}

\end{document}